\documentclass[acmtog]{acmart}
\acmSubmissionID{304}

\usepackage{booktabs} 
\usepackage{bbm} 
\usepackage{mathtools}
\usepackage{xfrac}

\usepackage[ruled]{algorithm2e} 

\SetAlFnt{\small}
\SetAlCapFnt{\small}
\SetAlCapNameFnt{\small}
\SetAlCapHSkip{0pt}

\usepackage{siunitx}
\DeclareMathOperator*{\argmax}{arg\,max}
\DeclareMathOperator*{\argmin}{arg\,min}

\newcommand{\twodots}{\mathinner {\ldotp \ldotp}}

\newcommand{\ftrad}{\ensuremath{\mathcal{F}^{\text{t}}}\xspace}
\newcommand{\fdiff}{\ensuremath{\mathcal{F}}\xspace}
\newcommand{\functTrad}{\ensuremath{f_t^{\text{t}}}\xspace}
\newcommand{\functDiff}{\ensuremath{f_t}\xspace}
\newcommand{\pixel}{\ensuremath{\mathrm{x}}\xspace}

\acmJournal{TOG}

\setcopyright{acmcopyright}\acmJournal{TOG}
\acmYear{2020}\acmVolume{39}\acmNumber{6}\acmArticle{262}\acmMonth{12} \acmDOI{10.1145/3414685.3417830}
\begin{document}
\title{Discovering Pattern Structure Using Differentiable Compositing}

\author{Pradyumna Reddy}
\affiliation{%
    \institution{University College London}
}

\author{Paul Guerrero}
\affiliation{%
    \institution{Adobe Research}
}

\author{Matt Fisher}
\affiliation{%
    \institution{Adobe Research}
}

\author{Wilmot Li}
\affiliation{%
    \institution{Adobe Research}
}

\author{Niloy J. Mitra}
\affiliation{%
    \institution{Adobe Research}
}
\affiliation{%
    \institution{University College London}
}

\begin{teaserfigure}
\centering
\includegraphics[width=\linewidth]{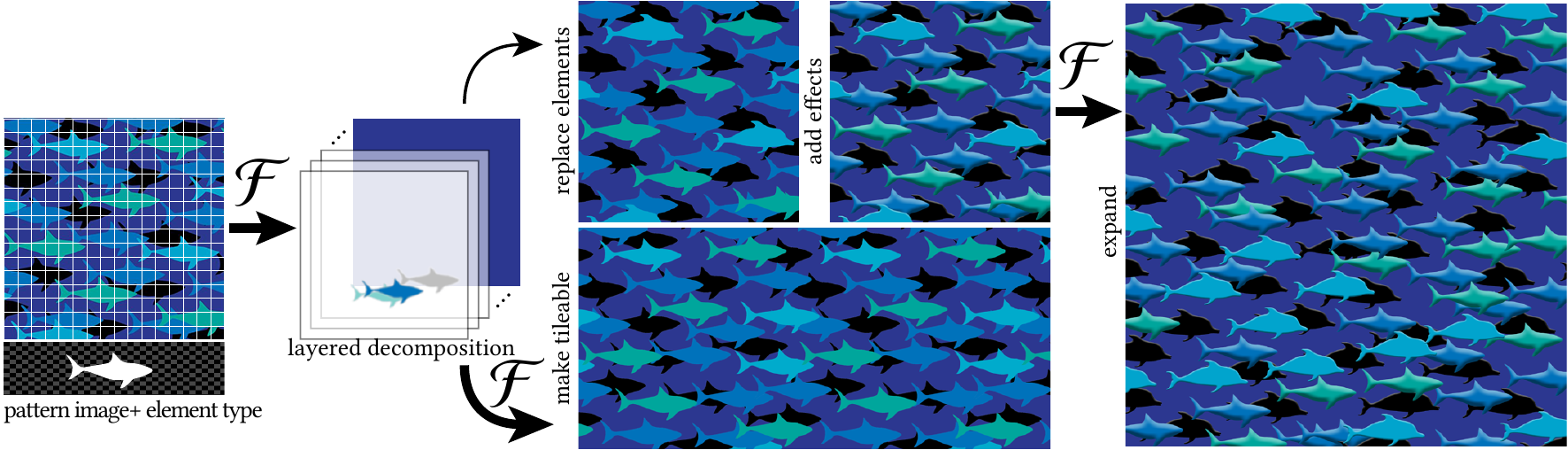}
\caption{
{Differentiable compositing.}
We present a differentiable function $\mathcal{F}$ to composite a set of discrete elements into a pattern image. This directly connects vector graphics to image-based losses (e.g., $L_2$ loss, style loss) and allows us to optimize discrete elements to minimize losses on the composited image. Minimizing an $L_2$ loss gives us a decomposition of an existing flat pattern image into a set of depth-ordered discrete elements that can be edited individually. Minimizing a style loss allows us to 
make a pattern tileable or expand a pattern image into a larger pattern composed of discrete elements.}
\label{fig:teaser}
\end{teaserfigure}

\begin{abstract}
Patterns, which are collections of elements arranged in regular or near-regular arrangements, are an important graphic art form and widely used due to their elegant simplicity and aesthetic appeal. 
When a pattern is encoded as a flat image without the underlying structure, manually editing the pattern is tedious and challenging as one has to both preserve  the individual element shapes and their original relative  arrangements. 
State-of-the-art deep learning frameworks that operate at the pixel level are unsuitable for manipulating such patterns. Specifically, these methods can easily disturb the shapes of the individual elements or their arrangement, and thus fail to preserve the latent structures of the input patterns. 
We present a novel \textit{differentiable compositing} operator using pattern elements and use it to discover structures, in the form of a layered representation of graphical objects, directly from raw pattern images. This operator allows us to adapt current deep learning based image methods to effectively handle patterns. 
We evaluate our method on a range of patterns and demonstrate superiority  in the context of pattern manipulations when compared against state-of-the-art
pixel- or point-based
alternatives. 
\end{abstract}

%
%


%
%


\maketitle

\section{Introduction}

Advances in deep learning, both in terms of network-based optimization as well as generative adversarial networks, have resulted in unprecedented advances in many classical image manipulation tasks. For example, deep learning is now the state-of-the-art in image denoising~\cite{pmlr-v80-lehtinen18a, Guo_2019_CVPR}, image stylization~\cite{gatys2016image, Karras_2019_CVPR, Jing2019NeuralStyleTransferReview}, image completion~\cite{Yu_2018_CVPR, Yu_2019_ICCV}, texture expansion~\cite{Zhou2018Non-StationaryFormat, Shaham_2019_ICCV} to name only a few. Generally, these methods directly operate on pixels and learn to optimize for image-space perceptual features or match kernel response statistics to produce compelling results.

Complementary to photographs and paintings, a dominant form of creative content is graphic patterns or simply patterns. Such patterns are created by artists as collections of discrete numbers of elements, often structured in regular or near-regular arrangements. The aesthetics of such patterns are dictated by both the arrangement of the discrete elements as well as the shape of the individual elements. 
Given a pattern, the user may may want to perform several manipulations: collectively edit the appearance of similar elements, replace the current set of elements with a different set, change the depth ordering of the elements, or redistribute the elements to match the style of a target pattern.

The common theme across the above applications is that they require the ability to synthesize the pattern according to some user-defined specifications while preserving the original pattern structure, i.e.,  the global arrangement of elements and the appearance of individual elements.
In most real-world examples, patterns are expressed as flat pixel images and such structures are not explicitly encoded. 

\begin{figure}[t!]
    \centering
    \includegraphics{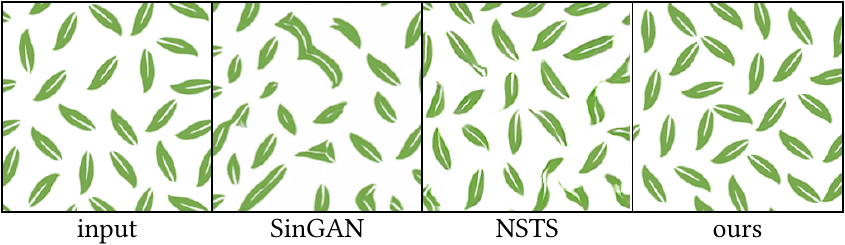}
    \caption{
    {Image-based DL methods can break elements in patterns.}
    Image-based pattern expansion methods like SinGAN~\cite{Shaham_2019_ICCV} or NSTS~\cite{Zhou2018Non-StationaryFormat} do not preserve element shapes. Our method is based on discrete elements and guarantees shape preservation. Here we show crops from expansions of the input pattern on the left.}
    \label{fig:image-based}
\end{figure}

Without direct access to the underlying pattern structure, limited options exist to effectively manipulate the patterns. 
One can treat an input pattern as an image, and use state-of-the-art image manipulation methods. These methods, being oblivious of the underlying elements and their arrangements, operate directly on the pixels and can easily destroy the original element shapes and their arrangements (see Figure~\ref{fig:image-based}). 
Alternately, one can manually select each individual element, arrange them into a layered structure, and then adjust the elements using image editing software. While such a workflow retains the shape of the respective elements, individually selecting and moving each element is both tedious and challenging especially when elements overlap, and the global arrangement of the elements must be manually preserved. 
%

In this paper, we develop a pattern manipulation framework that preserves \textit{both} the shape of the elements and their global arrangements. We achieve this by directly working with the individual elements, explicitly optimizing for element placements (i.e., locations, depth order, and orientations), while simultaneously assessing the quality of the arrangement of the manipulated pattern using image-domain statistics. 
Specifically, we optimize explicitly over the space of element placements while assessing the quality of the resultant pattern directly in the image domain using deep image features.
Our main contribution is a novel \textit{differentiable compositor}~(DC) that 
links, in a smooth and differentiable fashion, changes in the selection and placement of the the pattern elements to variations in features in the resultant pattern image.
For example, in Figure~\ref{fig:teaser}, starting from a flat pattern image with access to the different element types, the proposed DC can not only solve the inverse problem of disentangling the image into a layered representation of graphical objects (i.e., location, orientation, type, and depth order) along with the background image, but also be used to manipulate the patterns while keeping their original style. %
We further  generalize the proposed DC to handle multiple element types and increase robustness and convergence via a multiscale framework.

We evaluate our method on a variety of pattern designs of different complexity. Specifically, we compare against a state-of-the-art image space method~\cite{Zhou2018Non-StationaryFormat},
a point-based synthesis method~\cite{PointSyn19},
as well as a single-image similarity maximizing GAN~\cite{Shaham_2019_ICCV}. Our result shows the advantage of operating concurrently in the pixel domain and the structured elements domain via the proposed differentiable compositor. 
%

In summary, we introduce a novel differentiable compositing operator and use it to extend deep learning based image manipulation methods to directly operate on input patterns while preserving both the geometry of the individual elements and the image-domain similarity of the resultant patterns with respect to the original input. 
We will make our code publicly available upon publication.

\section{Related Work}

In our experiments, we focus on two applications of our differentiable compositing method: pattern decomposition, where a pattern image is decomposed into its constituent elements that con subsequently be edited, and pattern expansion, where a pattern image is used to synthesize a layout of discrete elements on a larger canvas that mimics the style of the original pattern. In this section, we briefly describe work relevant to these applications.


\subsection{Template matching}

Given a set of distinct elements that are used in a pattern, pattern decomposition can be approached as a template matching problem, where parameters of each element instance, such as its position and orientation, are found in the pattern image. A large body of prior work exists in template matching.
Here we present a subset that is relevant to our approach. Existing approaches differ mainly in their search strategies and similarity metrics. In the absence of gradients to guide the search for matches, the main employed strategies are variants of a grid search~\cite{korman2013fast, briechle2001template, wei2008fast, cheng2010repfinder}, which can get prohibitively expensive as the number of element parameters increases, and frequency domain matching~\cite{lewis1995fast, hel2005real, mattoccia2008fast}, which can only handle a limited set of element parameters. The similarity between the template and an image patch is measured with a variety of hand-crafted metrics in earlier methods. Ouyang et al.~\shortcite{ouyang2011performance} present an overview of search strategies and early similarity metrics. PatchMatch~\cite{barnes2009patchmatch} is probably the most established of the non-learning based image matching methods that uses a random search strategy followed by iterative growing of initial matches. We include a comparison to our differentiable approach in Section~\ref{sec:results}.

More recent methods use data-driven features~\cite{dekel2015best, talmi2017template} or data-driven similarity metrics~\cite{han2015matchnet, melekhov2016image,  hanif2019patch, luo2016efficient, wu2017deep, thewlis2016fully, bai2016exploiting, tang2016multi, rocco2017convolutional}
that typically learn a prior from large datasets. This results in improved robustness to conditions such as illumination changes and certain deformations, but may come at the cost of reduced generalization to templates that differ significantly from the examples in the training set. A Siamese network is typically trained with pairs of matched and unmatched patches to estimate the similarity, which can further be improved by carefully selecting training pairs~\cite{simo2015discriminative} or re-weighting matches based on their uniqueness~\cite{cheng2019qatm}.

Our method is orthogonal to these approaches. For the types of images we are investigating,
our method can provide gradients for the matching problem. These gradients can be used with any existing (differentiable) template similarity metric to make searching in a high-dimensional space of element parameters feasible.

Inverse procedural modeling systems~\cite{talton2011metropolis, vst2010inverse} share our goal of recovering a set of parameter values for a parametric model that generate a given output, although they use parametric formulations that are not differentiable, unlike ours, and typically work in different outputs domains (3D models or vector graphics).


\subsection{Image-based texture synthesis}
In this section we briefly describe image-based texture synthesis methods that are relevant to our pattern expansion application. Note that during synthesis, image-based methods do not distinguish between individual elements, making them less useful for down-stream editing.
Various deep learning based strategies have been successfully developed for this problem. These methods differ in the controls that they provide and there is a trade-off between the quality and the diversity of the generated output.


An early line of work synthesizes texture by optimally tiling patches from an exemplar over a an image region~\cite{efros2001image, hertzmann2001image, liang2001texSynth, kwatra2003graphcutTextures, barnes2009patchmatch, rosenberger2009layered}. The results can be used to expand exemplars into larger textures, fill holes, or to stylize and manipulate images.
In their influential work, Gatys et al.~\shortcite{gatys2016image} proposed a style loss for the synthesis of stylized images. They observed that the statistical correlation between certain deep features can be used as a measure of texture or stylistic similarity, where the correlation can be measured by the Gram matrix of the features computed with a pre-trained VGG-19 network~\cite{Simonyan2015VERYRECOGNITION}. Using this loss, texture or style can be transferred from one image to another via direct pixel-level optimization. 
Subsequently, Ulyanov et al.~\shortcite{Ulyanov2016TextureImages} proposed a variant of this approach that does not require optimization at inference time, by training a network to generate stylized images in a single feed-forward step. 
Since these methods have no notion of individual pattern elements, they cannot preserve the integrity of pattern elements,  when directly applied to pattern images (see Section~\ref{sec:results}). 
%
%

Zhou et al.~\shortcite{Zhou2018Non-StationaryFormat} proposed a framework to expand non-stationary textures to span larger canvases. The main idea is to combine a Gram matrix based texture similarity score, comparing random crops of the generated images to crops from the source image, and a GAN-based adversarial loss term. While the method works impressively on texture images, it also has no notion of individual elements, and can therefore not preserve the shape of the original elements. Further, the results lack any diversity in the output. 
%
%
%
Similar ideas~\cite{Shaham_2019_ICCV,Shocher_2019_ICCV}  have been proposed to learn  from a single image to generate a diverse set of similar looking images. Both of the approaches train a CNN in an adversarial setup to progressively generate images. While these techniques are able to synthesis diverse set of images, the generated images lose global structure, and can destroy characteristic arrangements in patterns. Also the generated images do not maintain the integrity of the elements in the training image. 

Differentiable renderers~\cite{liu2018paparazzi,liu2019soft,loubet2019reparameterizing} have been used to find deformations of 3D shapes that produce output images with a given style. These methods also rely on a differentiable image generation approach. However, in our work, we do not aim at moving triangles with small offsets, but rather at moving large elements with complex shapes and textures over relatively long distances to reach their target positions, while also optimizing for the number and types of elements.

\subsection{Point-based pattern synthesis}~
Several methods represent pattern elements as points in a high-dimensional parameter space, like the space of positions and orientations. Unlike the image-based representation, this representation takes into account the discrete nature of pattern elements, but loses the notion of element shape and appearance, both of which may influence the element layout in real-world patterns.
Some methods had success in synthesising textures with discrete elements~\cite{Landes2013ASynthesis, Hurtut2009Appearance-guidedExample, ma2011discrete, roveri2015example}, but are handcrafted for specific types of patterns and do not generalize well to other types of patterns, or require manually arranging the discrete elements into an exemplar texture.
More general methods~\cite{Zhou2012PointSpectrum, Oztireli2012AnalysisCorrelation, Heck2013BlueAliasing, Roveri2017GeneralCorrelations, Illian2008StatisticalPatterns} produce point patterns by matching spectral statistics, but cannot capture the local structure of elements due to the use of statistics. Patterns with discrete elements can also be manipulated efficiently by considering the shared geometric relationships between elements~\cite{GuerreroEtAl:PATEX:2016}, but require these relationships to be given, and cannot synthesize patterns from scratch.

More recently, deep learning methods for point cloud generation have been proposed~\cite{Li2018PointGAN, Sun2018PointGrow:Self-Attention, Achlioptas2017LearningClouds}, but these methods treat point clouds as 3D surfaces so they are not particularly suitable for point pattern synthesis out of the box. Other recent methods~\cite{leimkuehler2019DeepPointCorr, Muller2018NeuralSampling} deal with points from a sampling perspective, so the distribution of the points is taken into consideration. Recently, Tu et al.~\shortcite{tu2019point} present a method for expanding point patterns using a style loss. They represent points in the form of an image and use this exemplar as reference to optimize pixel values on larger canvas using style loss, histogram loss and correlation loss. While their results are compelling they do not account for the shape or appearance of the element. Also extending their method to deal with elements that are not interchangeable is not straight forward.

\begin{figure*}[t!]
    \centering
    \includegraphics[width=\linewidth]{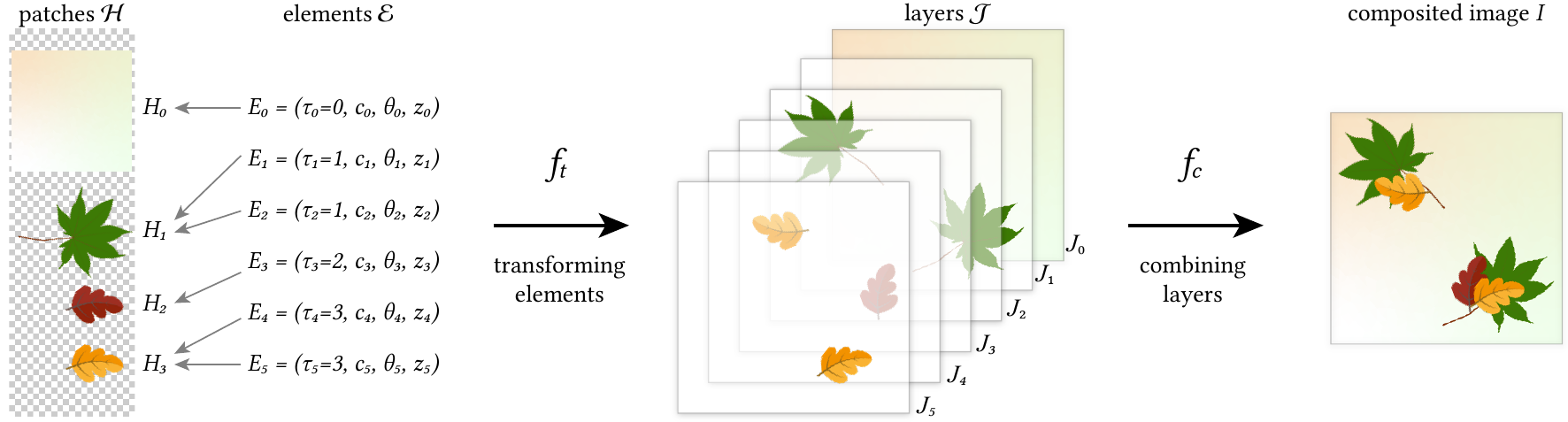}
    \caption{
    {Overview of the image compositing process. }
    Given a set of elements, each describing an instance of a small image patch at a position $c_i$, orientation $\theta_i$, and depth $z_i$, we first place each element in a separate image layer with the function $f_t$ and then combine the layers into the final image according to the element depth with function $f_c$, where layers with higher depth occlude layers with lower depth. We describe the traditional non-differentiable formulation of $f_t$ and $f_c$ in Section~\ref{sec:compositing_trad}, and our new differentiable formulation in Section~\ref{sec:compositing_diff}.}
    \label{fig:pipeline}
\end{figure*}

\section{Overview}
\label{sec:overview}
The prevalent approach to create illustrations and patterns is to \emph{composite} an image from a set of discrete elements. Specifically, a set of discrete elements, with each element having its own properties (i.e., type and its placement), are arranged on a canvas and combined into a single image. 
Our goal is to create a \emph{differentiable} compositing function for such discrete elements. Enabling the differentiability of the function, with respect to the element properties, allows us to leverage image-space loss on the composited image to directly drive the optimization of the discrete elements, both the choice of elements as well as their placements. This directly connects vectorized patterns to a range of losses available for images, such as L2 image loss or perceptual style loss, and in turn, enables vector graphics to benefit from advances in deep image processing.

In this work, we will focus on pattern images that are composed from a small set of atomic image patches that are instanced multiple times and arranged, possibly as overlapping layers, over a background canvas to form a pattern. We call the instances of these patches pattern \emph{elements}. Each element is defined by a set of parameters, such as its position, orientation, and depth.

A compositing function that takes these elements and outputs a composited image is illustrated in Figure~\ref{fig:pipeline} and can be described in two main steps: first, a \emph{layer} is created for each element that has the same size as the output image and contains the image patch corresponding to the element, at the position and orientation given by the element parameters; second, the layers are combined according to the depth given by the element parameters, where elements in layers with higher depth occlude layers with lower depth.

In Section~\ref{sec:compositing_trad}, we present the traditional formulation $\ftrad$ of this compositing function, as it is often used in software such as Photoshop, Illustrator, and Powerpoint. Due to discrete quantities such as the number of elements and their visibility, this formulation is not differentiable. In Section~\ref{sec:compositing_diff}, we show, as our core contribution, that this formulation can be generalized as a differentiable image composition function $\fdiff$. In Section~\ref{sec:optimization}, we describe our approach to optimize image compositions w.r.t. the element parameters using our differentiable formulation (see Table~\ref{tab:notations} for notations).




\begin{table}[b!]
\caption{Notation table. }
\label{tab:notations}
\begin{tabular}{r|p{0.6\columnwidth}}
symbol                      & description                   \\ \hline \hline
$I$                         & RGB image                         \\
$\mathbf{x}$                & image location \\
$\mathcal{E}\coloneqq\{E_0 \twodots E_n\}$               & set of elements      \\
$\fdiff$                         & (differentiable) compositing function          \\
$\mathcal{H}\coloneqq\{H_0 \twodots H_m\}$               & library of image patches      \\
$E_i \coloneqq (\tau_i, c_i, \theta_i, z_i)$                       & element properties            \\ 
$\tau_i$ &  type of element $i$ \\
$c_i$ &  center location of element $i$ \\
$\theta_i$ &  orientation of element $i$ \\
$z_i$ &  layer depth of element $i$ \\
$f_t$                  & element transformation function \\
$f_c$                  & element compositing function \\
$\mathcal{J}\coloneqq\{J_0 \twodots J_n\}$               & layers                        \\
$T_i$                       & element transformation matrix \\
$\kappa$                    & interpolation kernel        \\
$v_i(x)$                    & visibility function of layer $i$        \\
$t_i = (t^0_i \twodots t^m_i)$ & type probability vector of element $i$
\end{tabular}
\end{table}


\section{Traditional Image Compositing}
\label{sec:compositing_trad}
In a forward process, we create an RGB image $I^{\text{t}}$ by compositing a given set of $n$ discrete elements $\mathcal{E}^{\text{t}} \coloneqq \{E_0^{\text{t}} \twodots E_n^{\text{t}}\}$
with a function $\ftrad$ to produce image value at 2D image coordinate $\pixel$ as, 
\begin{equation}
    I^{\text{t}}(\pixel) \coloneqq \ftrad(\pixel, \mathcal{E}^{\text{t}}).
\end{equation}
Each element $E_i^{\text{t}}$ is defined as a translated and rotated instance of an $RGBM$ image patch $H_j$ selected from a small library of image patches $\mathcal{H} = \{H_1 \twodots H_m\}$, where $M$ is a binary coverage mask that defines the region of the image patch covered by the element.
An image patch represents the appearance of an element and is typically much smaller than the image $I$. Each element consists of a property tuple $E_i^{\text{t}} \coloneqq (\tau_i, c_i, \theta_i, z_i)$, where $\tau_i \in \{1 \twodots m\}$ is an index into $\mathcal{H}$ that denotes the element type, $c_i \in \mathbb{R}^2$ denotes the coordinates of the element center, $\theta_i \in[0,2\pi)$ denotes the orientation of the element, and $z_i \in \mathbb{Z}^+$ defines the occlusion order of the elements. In regions of overlap, elements with higher $z$ occlude elements with lower $z$. In the case of patterns, we typically have a small set of element types (e.g., $m \le 5$) each of which appears multiple times in the pattern image (i.e., $m \ll n$). 
We treat the background  $H_0$ as a special element with a fixed set of parameters $E^t_0 \coloneqq (\tau_0=0, c_0=(0,0), \theta_0=0, z_0=0)$.

%

Given this collection of ingredients, we create an image in two steps.  First, the function $f_t^\text{t}$ transforms the image patch of each element according to its property vector, and then re-samples the transformed image patch at the pixels of the image $I$ to obtain one image layer $J_i^{\text{t}}$ per element. Second, the function $f_c^\text{t}$ combines the occlusion-ordered individual layers into a single image:
\begin{equation}
    I^{\text{t}}(\pixel) \coloneqq \ftrad(\pixel, \mathcal{E}^{\text{t}}) = f_c^\text{t} \Big( \big\{f_t^{\text{t}}(\pixel, E_i^{\text{t}})\big\}_{i=0:n} \Big).
\end{equation}

\paragraph{Transforming elements} In the first step, we rasterize each element on its separate image layer, resulting in a set of layers $\mathcal{J}^t = \{J_0^{\text{t}} \twodots J_n^{\text{t}}\}$ as $J_i^{\text{t}}(\pixel) = f_t^\text{t}(\pixel, E_i^{\text{t}})$.
We initialize the $RGBM$ channels of each layer to zero and then place the image patch of the corresponding element by translating it to its center position $c_i$ and orienting it according to $\theta_i$. For layer $i$, we transform the image coordinates \pixel into the local coordinate frame of the element using the element's inverse transform, and then sample the image patch of the element with the local coordinates as:
\begin{equation}
   \functTrad(\pixel, E_i^{\text{t}}) \coloneqq   H_{\tau_i} \big(\lfloor R_{\theta_i}^{-1} (\pixel-c_i) \rceil\big),
  \label{eqn:trad_rounding} 
\end{equation}
where $R_\theta$ denotes a $2 \times 2$ matrix encoding rotation by  angle $\theta$ and $\lfloor \rceil$ indicates rounding to the nearest pixel location.
%

\paragraph{Combining occlusion-ordered element layers} 
In the second step, we combine the layers $\mathcal{J}$ to get the composited image $I^{\text{t}}(\pixel) = f_c^\text{t}(\{J_i^{\text{t}}(\pixel)\}_{i=0:n})$ by stacking layers on top of each other according to the occlusion order of their elements, where layers with a higher $z$-value occlude layers with with lower $z$.
This gives the following definition for the composited image:
\begin{equation}
I^{\text{t}}(\pixel) = \sum_{i=0}^n J_i^{\text{t}}(\pixel) v_i(\pixel)
\end{equation}
where $v^{\text{t}}_i(\pixel) \in \{0,1\}$ is the visibility of layer $i$ at each image location $\pixel$.
Only the layer with non-zero mask $M$ and highest $z$-value is visible at each image location, i.e.,
\begin{equation}
    v^{\text{t}}_i(\pixel) =
    \left\{\begin{array}{ll}
        1 & \text{if } i \in \argmax_j z_j M^{\text{t}}_j(\pixel)\\
        0 & \text{otherwise},
        \end{array}\right.
\label{eqn:trad_visibility}        
\end{equation}
where $M^{\text{t}}_j$ is the coverage mask of the layer $J^t_j$.
 

\section{Differentiable Image Compositing}
\label{sec:compositing_diff}
Given a composited image, the inverse problem is to recover a set of elements, both their count and individual properties.
Solving this problem requires differentiating the compositing function with respect to the (unknown) element properties. However, doing so with the traditional compositing function, introduced above, is difficult due to several factors:
(i)~the nearest neighbor rounding, see Equation~\ref{eqn:trad_rounding},
prevents taking gradients with respect to element location and/or orientation; (ii)~the discrete visibility $v^\text{t}_i$, see Equation~\ref{eqn:trad_visibility}, prevents taking gradients with respect to the element occlusion order; (iii)~the count $n$ of elements and (iv)~their types $\tau_i$ 
are discrete variables preventing us from taking their derivatives. 

We introduce a differentiable image compositing function $\fdiff$ that addresses these challenges by generalizing the traditional compositing function $\ftrad$. In particular, we replace the nearest neighbor rounding with a differentiable interpolation kernel for sampling image patches, and convert the discrete variables in $\ftrad$ into soft, continuous variables in $\fdiff$, and use a multi-resolution pyramid to smooth the composite spatially. While the basic image formation process remains similar, these generalizations require some important adjustments to the compositing formulation.

%
%

Similar to before, we create an RGB image $I$ by compositing a set of $n$ discrete elements $\mathcal{E} \coloneqq \{E_0 \twodots E_n\}$ with a differentiable function $\fdiff$ as, 
\begin{equation}
    I(\pixel) \coloneqq \fdiff(\pixel, \mathcal{E}),
\end{equation}
where \pixel are 2D image coordinates. We also assume availability of a library of $RGBM$ image patches $\mathcal{H} \coloneqq \{H_1 \twodots, H_m\}$. Each element is given a tuple of variables $E_i^{\text{t}} \coloneqq (t_i, c_i, \theta_i, z_i)$, where
$t_i = (t^1_i \twodots, t^m_i)$ softens the element type $\tau_i$ into a vector of logits $t^j_i$ of the probability that element $E_i$ is of type $j$,
$c_i \in \mathbb{R}^2$ denotes the coordinates of the element center, $\theta_i \in \mathbb{R}$ 
denotes the orientation of the element in radians, and $z_i \in \mathbb{R}$ defines the occlusion order of the elements, where elements with higher $z$ occlude elements with lower $z$. The background $H_0$ is represented with the element $E_0 \coloneqq (c_0=(0,0), \theta_0=0, z_0)$ that does not have type logits $t_0$ and is placed at a fixed depth $z_0$. Unlike in traditional compositing, the depth $z_i$ of any element can be smaller than the background depth $z_0$, making it invisible in the final composite, due to being fully occluded by the background. This provides a simple mechanism to make the number of elements differentiable without introducing additional parameters. We set $n$ to a large number (between $100$ and $1024$ in our experiments), and remove any elements in the final composite that end up hidden below the background (i.e., $z_i < z_0$).

Similar to the earlier formulation, we define the composited image in two steps. First, the function $f_t$ transforms the image patches of each element according to the element properties, and then re-samples the transformed image patch at the pixels of the image $I$ to obtain one image layer $J_i$ per element. Second, the function $f_c$ combines the layers according to the occlusion order into a single image:
\begin{equation}
    I(\pixel) := \fdiff (\pixel, \mathcal{E}) = f_c \Big( \big\{ \functDiff(\pixel, E_i)\big\}_{i=0:n} \Big).
\end{equation}
Our goal is to define a function $\fdiff$ that is differentiable with respect to the properties $E_i$ of all elements. Later, we show how the differentiable function   allows computing the gradients $\frac{\partial L(I)}{\partial E_i}$ of any differentiable image loss $L(I)$ with respect to the properties $E_i$. 

\paragraph{Transforming elements} In the first step, we place each element on a separate image layer, resulting in a set of layers $\mathcal{J} := \{J_1 \twodots J_n\}$ as $J_i(\pixel) :=  \functDiff(\pixel, E_i)$.
We set ${J}_0$ with depth $z_0=\gamma$ with a (unknown) constant color $J_0(\pixel) = b$ and alpha = 1. 
We initialize the $RGBM$ channels of each layer to zero and then place the image patch of the corresponding element by translating it to its center position $c_i$ and orienting it according to $\theta_i$. 
For layer $i$, we transform the image coordinates \pixel into the local coordinate frame of the element using the element's inverse transform, and then sample the image patch of the element with the local coordinates as:
\begin{equation}
      h_{\tau_i} \big(  R_{\theta_i}^{-1} (\pixel-c_i)  \big),
  \label{eqn:rounding} 
\end{equation}
where $R_\theta$ denotes a $2 \times 2$ matrix encoding rotation by  angle $\theta$ and  $h_j$ is an interpolation of image patch $H_j$ to give a continuous function:
\begin{equation}
    h_j(\pixel) = \sum_k (H_j)_k\ \kappa(\pixel - \pixel_k),
\end{equation}
where $(H_j)_k$ is pixel $k$ of the image patch $H_j$, $\pixel_k$ are the coordinates of the same pixel, and $\kappa$ is a differentiable interpolation kernel. 

We create a differentiable function $\functDiff(\pixel, E_i)$ using the expected value over type probabilities:
\begin{equation}
\label{eq:diff_type}
J_i(\pixel) := \functDiff(\pixel, E_i) = \frac{1}{\sum_{k=1}^m e^{t^k_i}} \sum_{j=1}^m e^{t^j_i}\ h_j \big( R_{\theta_i}^{-1} (\pixel-c_i) \big),
\end{equation}
where the softmax $e^{t^j_i} /\ \sum_{k=1}^m e^{t^k_i}$
over type logits define the type probabilities.
In this formulation, the gradients with respect to the element type probabilities $t^j_i$ are well-defined. Note that the expected value is over all four $RGBM$ channels of the interpolated image patch $h_j$, and thus the coverage mask $M_i$ can take on non-binary values,
effectively giving us a map of the coverage probability in layer $J_i$. The background layer is computed as $J_0(\pixel) := h_0 \big( R_{\theta_0}^{-1} (\pixel-c_0) \big)$.

\begin{figure}[t]
    \centering
    \includegraphics[width=\linewidth]{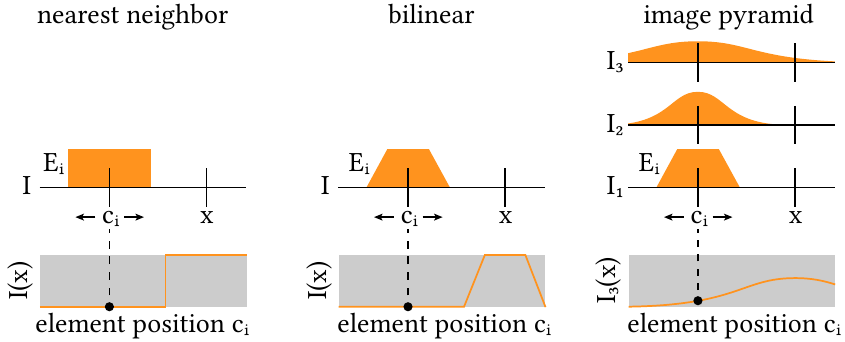}
    \caption{
    {Differentiable element position.}
    Nearest-neighbor sampling of an element results in an image $I$ that is not differentiable w.r.t. the element position $c_i$. Using bilinear sampling improves differentiability, but gradients at $I(\pixel)$ only have a small region of influence. Using an image pyramid further improves differetiability and increases the region of influence.}
    \label{fig:diff_position}
\end{figure}

%
Empirically, we found that the choice of interpolation kernel $\kappa$ is important to get good gradients with respect to the location and orientation of the elements. We choose a bilinear interpolation kernel (cf., \cite{jaderberg2015spatial}):
\begin{equation}
    \kappa(\pixel) = \frac{1}{r_x r_y} \max(0, r_x-|\pixel_x|) \max(0, r_y-|\pixel_y|),
\end{equation}
where $r_x$ and $r_y$ is the sample spacing along the $x$ and $y$ axes of the image patch $H_i$, which makes $\functDiff$ piecewise differentiable with respect to the position and orientation of the elements, see Figure~\ref{fig:diff_position} for an illustration.

\begin{figure}[t]
    \centering
    \includegraphics[width=\linewidth]{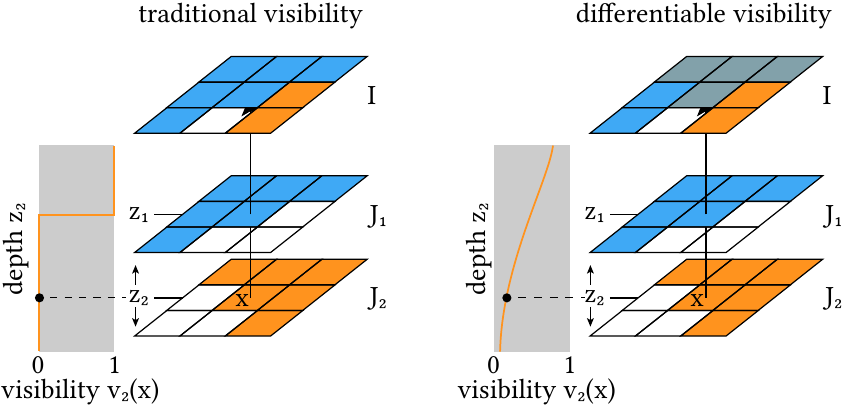}
    \caption{
    {Differentiable visibility.} The traditional visibility function is not differentiable due to $C^0$ discontinuities at disocclusions. In our differentiable formulation, each layer contributes to the final image with a weight that decreases exponentially with distance from the top-most layer.}
    \label{fig:diff_visibility}
\end{figure}

\paragraph{Combining element layers} 
In the second step, we combine the layers $\mathcal{J}$ to get the composited image $I(\pixel) = f_c(\{J_i(\pixel)\}_{i=0:n})$. Layers are stacked on top of each other according to the occlusion order of their elements, where layers with a higher $z$-value occlude layers with lower $z$. The final composited image is defined as:
\begin{equation}
I(\pixel) := f_c(\{J_i(\pixel)\}_i) = \sum_{i=0}^n J_i(\pixel) v_i(\pixel),
\end{equation}
where the visibility $v_i(\pixel)$ determines which layer is visible at each image location $\pixel$. We make this visibility differentiable with respect to the last remaining element property: the z-value $z_i$.
We use a differentiable version of the visibility using a softmax function as:
\begin{equation}
v_i(\pixel) := {e^{z_i} M_i(\pixel) } / {\sum_{k=0}^n{e^{z_k} M_k(\pixel) }}.
\end{equation}
In this softer version of the visibility, all elements with non-zero alpha have some contribution to the image, even if they are occluded, but their contribution decreases exponentially with distance from the top-most element, see Figure~\ref{fig:diff_visibility} for an illustration. Recall from Eq.\ref{eq:diff_type} that $M_i$ is the coverage probability in layer $J_i$. By multiplying with $M_i$, regions of the layer that are likely to be occupied are placed near the depth $z_i$ of the element, while regions that are less likely to be occupied are placed at lower depth, making them less likely to occlude other elements.
Due to our continuous formulation, variables that are discrete in the traditional formulation, like the probabilities for element types, can take on non-integer values. This results in artifacts like ghosted semi-transparent elements. In Section~\ref{sec:optimization}, we describe how we project the continuous variables to integer values.

\paragraph{Multi-resolution pyramid}
The bilinear interpolation kernel makes the composited image $I$ differentiable w.r.t. the element positions $c_i$ and orientations $\theta_i$. However, the gradients $\frac{\partial I}{\partial c_i}$ and $\frac{\partial I}{\partial \theta_i}$ have a small \emph{region of influence}: they are non-zero only at image locations $x$ that are close to the coverage mask of the element, at a distance smaller than the bilinear kernel's radius $r$, see Figure~\ref{fig:diff_position} for an illustration. This adversely affects convergence if the initial pose of an element is not already very close to its target region, since gradients of the target region w.r.t the element parameters are zero. Additionally, high spatial frequencies in the image $I$ result in unstable gradients. To improve convergence, we compute the gradients at each level of a multi-resolution pyramid:
\begin{align}
    I_1 &= I, \text{and} \nonumber \\
    I_{k+1} &= I_{k} \circledast G(0,\sigma_k),
\end{align}
where $G(0,\sigma_k)$ is a Gaussian kernel with standard deviation $\sigma_k=r 2^{k-1}$, and $r$ is the radius of the bilinear kernel. To improve performance, we successively down-sample the image to half its resolution in each iteration. At coarser levels, the gradients $\frac{\partial I_k}{\partial c_i}$ and $\frac{\partial I_k}{\partial \theta_i}$ are both more stable and have a larger region of influence, at the cost of spatial accuracy.
%
%
In our experiments, we use gradients from four pyramid levels ($i=1\twodots4$).

\paragraph{Differentiable element colors}
It is straight-forward to add other element properties to our differentiable formulation. As an example, we can add an element color $o_i \in \mathbb{R}^3$ by simply re-defining the element layers in Eq.~\ref{eq:diff_type} as 
$J'_i(x) = J_i(x) \cdot \nu(o_i)$, where $\nu$ is a leaky hard sigmoid, and $\cdot$ denotes a multiplication of each RGB channel in $J_i$ with the corresponding scalar color value in $\nu(o_i)$. We chose a leaky hard sigmoid $\nu(o_i)=\max(\min(o_i,\ 0.001 o_i + 0.99), 0.001 o_i)$ instead of the regular sigmoid to make it easer for the optimization to reach fully saturated colors. We will show several composited images with optimized color parameters $o_i$ in our results.

\section{Optimizing Image Compositions}
\label{sec:optimization}

Given our differentiable formulation of image compositing, we can optimize the element parameters to minimize any loss function $L(I)$ defined on the composited image $I$. In our experiments, we use the Adam~\cite{kingma2014adam} optimizer, with a learning rate of $\num{1e-6}$ and parameters $(\beta_1, \beta_2) = (0.9, 0.9)$. Since each parameter has a different value range, we adjust the learning rates for each type of parameter. Empirically, we multiply the learning rate for parameters $(t_i, c_i, \theta_i, z_i, o_i)$ with the factors $(1, 0.01, 2.25, 0.0016, 0.0025)$ in all our experiments.
A straight-forward optimization with a random initialization is difficult due to several factors: (i) as described earlier, gradients have a limited region of influence, which limits the basin of attraction for the optimal position and orientation of an element, and (ii) multiple elements may be attracted to the same optimum in position and orientation, potentially giving us duplicate elements.
To overcome these problems, we carefully initialize the element parameters, and periodically seed additional elements and remove elements that are no longer visible or identified as duplicates.

\paragraph{Initialization}
We start by choosing an estimated upper bound for the number of elements we are going to need, between $100$ and $1024$ elements in our experiments. The centers $c_i$ of these elements are initialized in a regular grid over the image canvas, with orientation $\theta_i = 0$. The element types logits are initialized to $t^j_i = 1$ for all types $j$ and the depth of the background and the other elements is initialized empirically to $z_0=3.3$ and $z_i = 9$, respectively. The grid initialization for element positions makes it less likely to miss the basin of attraction of a position optimum. A grid initialization for both positions and orientations would result in too many elements, so we handle missing orientations when seeding additional elements during the optimization.

\paragraph{Removing elements and seeding additional elements}
We alternate between removing and re-seeding elements periodically during the optimization. Elements are removed at iterations $4$k, $12$k, and every $8$k iterations thereafter, and re-seeded three times at iterations $8$k, $16$k, and $24$k.
We remove elements that are no longer visible because they are below the background $z_i < z_0$, and duplicate elements. 
Specifically, we identify pairs of duplicates elements as having the same type $\tau_i = \argmax(t_i)$ and a distance $\|c_i - c_j\|_2 < 0.5 \min\big(l(M_i), l(M_j)\big)$ where $l(M)$ is the largest extent of an element's mask along the $x$ and $y$ axis. We remove the element with smaller depth in such a duplicate pair.

In the re-seeding steps, we sample the space of element orientations with a regular grid. For each existing element, we place three additional copies with the same parameters, but with orientation offsets of $90$, $180$, and $270$ degrees. Note that, after subsequent optimization, wrongly oriented elements end up below the $H_0$ layer, and hence are later removed. This allows us to escape local minima in the space of element orientations, which are especially pronounced if elements are close to rotationally symmetric.
Additionally, we place a new grid of elements, with the same parameters as in the initialization. These additional elements allow convergence to position optima that were missed with the previous set of elements.

\paragraph{Discretization}
After optimization, we perform a final \emph{discretization} step that removes any residual transparencies or ghosting introduced by our continuous parameters. We generate a discretized image by running our optimized element parameters through the traditional compositing pipeline, using discrete element type probabilities $\tau_i = \argmax(t_i)$.



\section{Results and Discussion}
\label{sec:results}

Pattern images are ubiquitous on the web, but are difficult to edit for several reasons. The constituent elements of a pattern are difficult to extract manually from the image, due to their number and the frequency of occlusions. Additionally, editing individual elements may compromise the style of a pattern, and restoring the style might require (manually) modifying all pattern elements.
Below we evaluate our differentiable compositing function on two applications that are targeted at simplifying the editing workflow for pattern images: in Section~\ref{sec:results_decomp}, we experiment with decomposing patterns images into a set of discrete elements that are editable individually, and in Section~\ref{sec:results_expansion}, we experiment with pattern expansion, where, given a pattern image, a new pattern with discrete elements and the same style as the input image is synthesized. Each of the two applications corresponds to a minimization of a different loss function. Additionally, we show the versatility of our differentiable compositing function with a few additional applications in Section~\ref{sec:results_additional}.

\subsection{Pattern Decomposition}
\label{sec:results_decomp}

In this application, our aim is to decompose a given pattern image $A$ into a set of elements $\mathcal{E}$. The image composited from these elements $I = \mathcal{F}(\mathcal{E})$ should reconstruct the input image $A$ as closely as possible. We use the $L_2$ distance to measure the decomposition error:
\begin{equation}
L_d(A, I) := \frac{1}{P} \sum_{p=1}^P \|A(x_p)-I(x_p)\|^2_2,
\end{equation}
where the sum is over pixels and $P$ is the number of pixels in the image. The optimal elements $\mathcal{E}^*$ are found by minimizing this loss:
\begin{equation}
\mathcal{E}^* := \argmin_{\mathcal{E}} L_d(A, \mathcal{F}(\mathcal{E})).
\end{equation}
Note that the compositing function $\mathcal{F}$ also depends implicitly on the set of image patches $\mathcal{H}$, corresponding to the different types of elements in the pattern. These are are selected by the user from the input image in the simple pattern decomposition workflow described below.

\begin{figure}[b!]
    \centering
    \includegraphics[width=\linewidth]{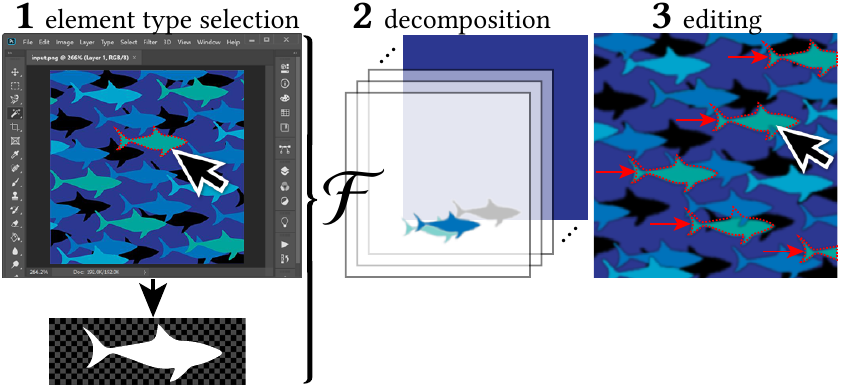}
    \caption{
    {Pattern decomposition workflow.} Starting from a pattern image, the user selects one example of each element type. Here, we only select a single element since we optimize over element colors. The input image and the element types form the input to our differentiable compositor, resulting in discrete elements that we can edit directly or use in down-stream tasks.}
    \label{fig:workflow}
\end{figure}

\paragraph{Workflow}
A typical workflow is illustrated in Figure~\ref{fig:workflow}. The user starts by selecting one example of each element type in the pattern image. This can usually be done efficiently using existing methods~\cite{cheng2010repfinder,rother2004grabcut} or image editing software such as Photoshop or GIMP and is significantly easier than selecting all instances of each element type in the pattern image. Together with the input image, these form the input of our optimization, which finds the set of discrete elements in the pattern. The discrete elements can now be edited by the user. We show a few examples of possible edits in Figure~\ref{fig:teaser}. See the accompanying video for additional edits. 

\begin{figure*}
    \centering
    \includegraphics[width=\textwidth]{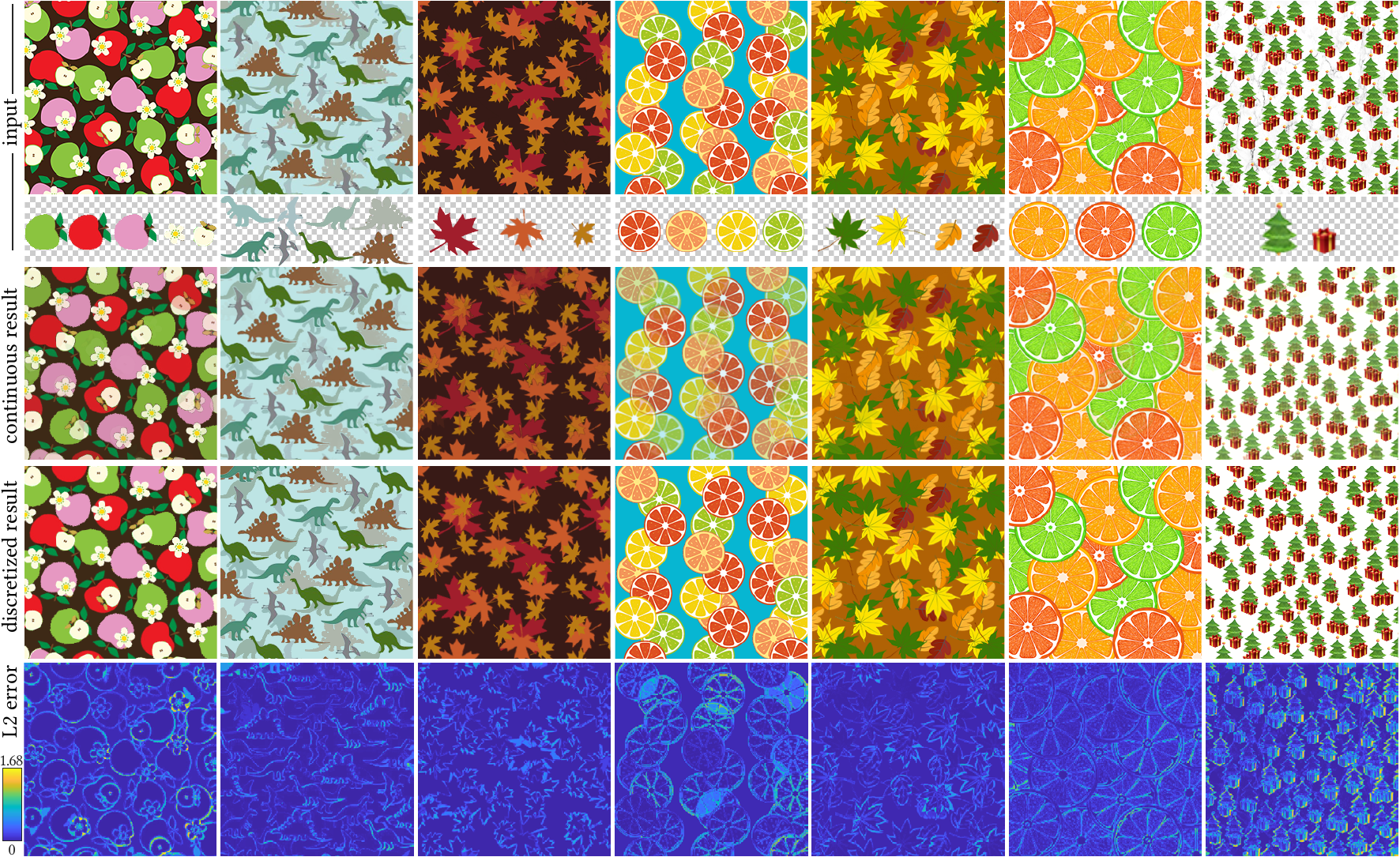}
    \caption{
    {Inverse Pattern decompositions.} Given the flat pattern images and the elements in the top row, we optimize for elements that reconstruct the pattern image using our differentiable compositing function. The result of the continuous optimization is shown in the second row, and the discretized result (inferred layering information is not shown here) in the third row. In the last row, we show the $L_2$ error between the discretized result and the input image for colors in $[0,1]^3$. The majority of the errors are evident near the boundaries due to small position inaccuracies of the elements.}
    \label{fig:results_decomposition}
\end{figure*}

\paragraph{Results}
Pattern decomposition results are shown in Figure~\ref{fig:results_decomposition}. We apply the pattern decomposition workflow to  pattern images downloaded from the web and two manually created pattern images (patterns \#4 and \#7). We only optimize over necessary parameters in each example. These include all parameters in examples \#1, \#3 and \#5, and all parameters except the orientation in the other examples.

The pattern image and the element types extracted by the user are given in the first row, and in rows 2 and 3, we show both the output of our optimization before the discretization step described in Section~\ref{sec:optimization} (\emph{continous result}), and after the discretization (\emph{discretized result}).
Please note the slight transparency or ghosting in some elements of the continuous result. This is due to the soft approximations of the visibility and element type we use in our differentiable composting function. Running our optimized element parameters through the non-differentiable compositor removes these artifacts and gives us the discretized results.
A map of the $L_2$ error between the discretized result and the input image is given in the last row. Note that our composited image closely resembles the original, with errors concentrated at boundaries, due to slight inaccuracies in the optimized element parameters. Apart from these inaccuracies, errors occur only very sparsely; for example, orientation errors in cases where objects are close to rotationally symmetric (column 5), or errors in the depth or type of an element if the shapes and colors of different element types are similar. We believe that most of these errors could be avoided with a larger number of optimization steps, or an improved optimization setup, such as a learning rate schedule.
Overall, we can see that our optimization can successfully find the type, position, orientation, and layering of elements that form a given pattern image, even in the presence of severe occlusions between elements.

\begin{figure}
    \centering
    \includegraphics[width=\linewidth]{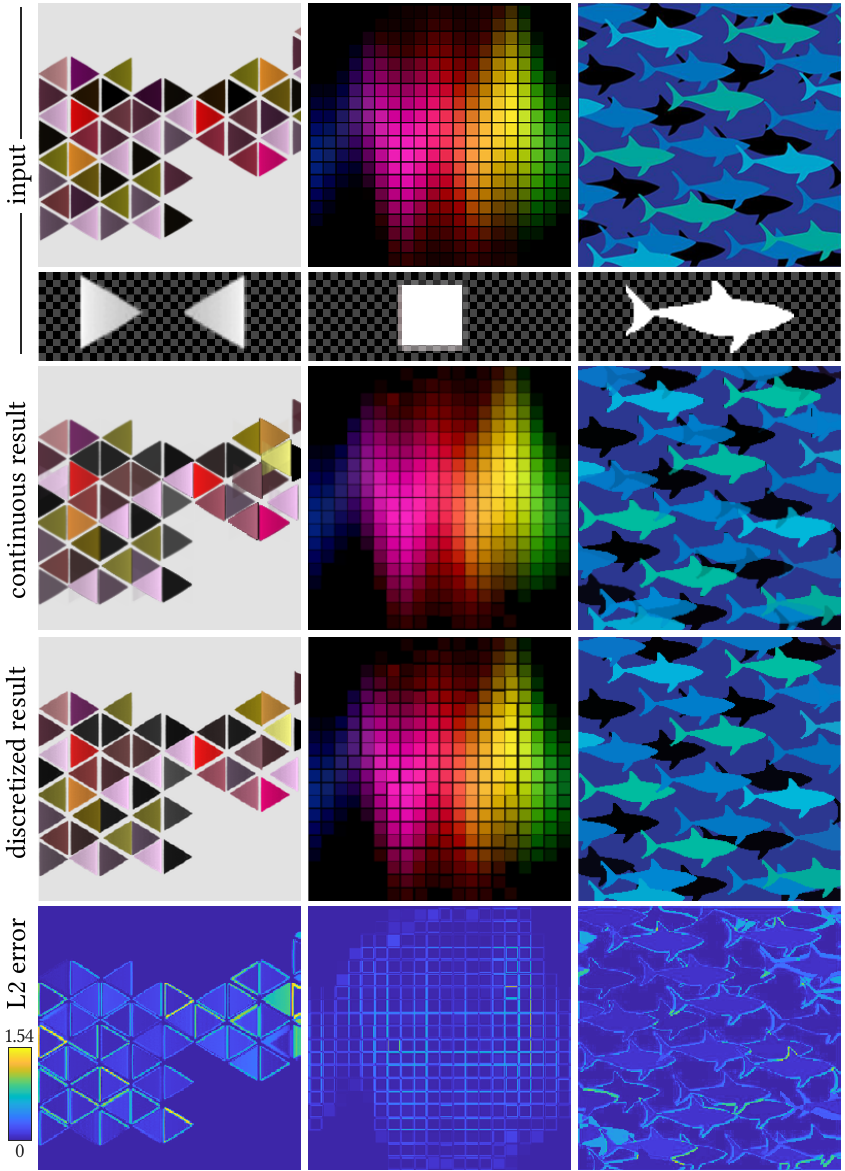}
    \caption{
    {Additional inverse pattern decompositions}. In these decompositions, we also optimize for the color (scaling) of the elements.}
    \label{fig:results_decomposition_additional}
\end{figure}

In Figure~\ref{fig:results_decomposition_additional}, we show additional decomposition results where we also optimize over the color $o_i$ of each element. All pattern images were downloaded from the web. The first pattern shows a simple element in large number of color variations that would be infeasible to handle without optimizing for colors, since we would need one element type per color. In the second pattern image, we show an example of elements with some amount of texture. Since we multiply the layer of each element with its color parameter, we can effectively also handle some amount of texture when optimizing for colors.
Finally, in the third column, we optimize for color in a pattern with  complex layering and occlusions.
As we can see in the last row, the performance of our optimization with color is on par with the optimization without color, showing that we can successfully optimize for element color as well.

\paragraph{Baseline comparison}
We compare our method to two baselines that do not benefit from our gradients. A brute-force grid search over element parameters and PatchMatch~\cite{barnes2009patchmatch}. We describe each baseline shortly before presenting comparison results.

As a first baseline, a brute force grid search over element parameters is solved greedily, one element at a time. We create a grid over all element parameters that are relevant for a given input image, except for the depth, which is handled with a greedy approach. We set the resolution of this grid to $(m, 128^2, 36, 3^3)$ for the parameters $E'_k = (\tau_k, c_k, \theta_k, o_k)$, where $m$ is the number of element types.
As a distance measure between an element and the input image, we define an $L_2$ distance that is normalized by the occupancy masks $M$ of both the element and the input image:
\begin{equation}
L_m(A, J_k') = \frac{\sum_{p=1}^P M_A(x_p) M_{J_k'}(x_p)\ \|J_k'(x_p) - A(x_p)\|_2^2}{\sum_{p=1}^P M_A(x_p) M_{J_k'}(x_p)},
\end{equation}
where $A$ is the input image, $J_k'(\pixel) = f_t^\text{t}(\pixel, E_k')$ is a layer containing the element transformed to grid location $k$ (see Equation~\ref{eqn:trad_rounding} for details), and $E_k'$ are the element parameters at a grid location $k$. This distance effectively ignores background pixels that are not occupied by the element. The image starts with full occupancy, but we successively remove matched regions from its occupancy mask, which are then ignored in subsequent matches. We first compute $L_m$ at each grid location and greedily pick the best match. Then, we remove the matched region from the image by subtracting the occupancy mask of the matched element from the occupancy mask of the image and update $L_m$ at all affected grid locations. This procedure is iterated until $L_m$ falls below a threshold. We \textit{manually} select the best threshold for each image.
In practice, we compute the sum only over the non-zero pixels of the element's occupancy mask and ignore matches where the number of non-zero pixels in the product of the occupancy masks is smaller than a threshold. In our experiments, we set this threshold to $10\%$ of the total number of non-zero pixels in the element's occupancy mask.

\begin{figure}
    \centering
    \includegraphics[width=\linewidth]{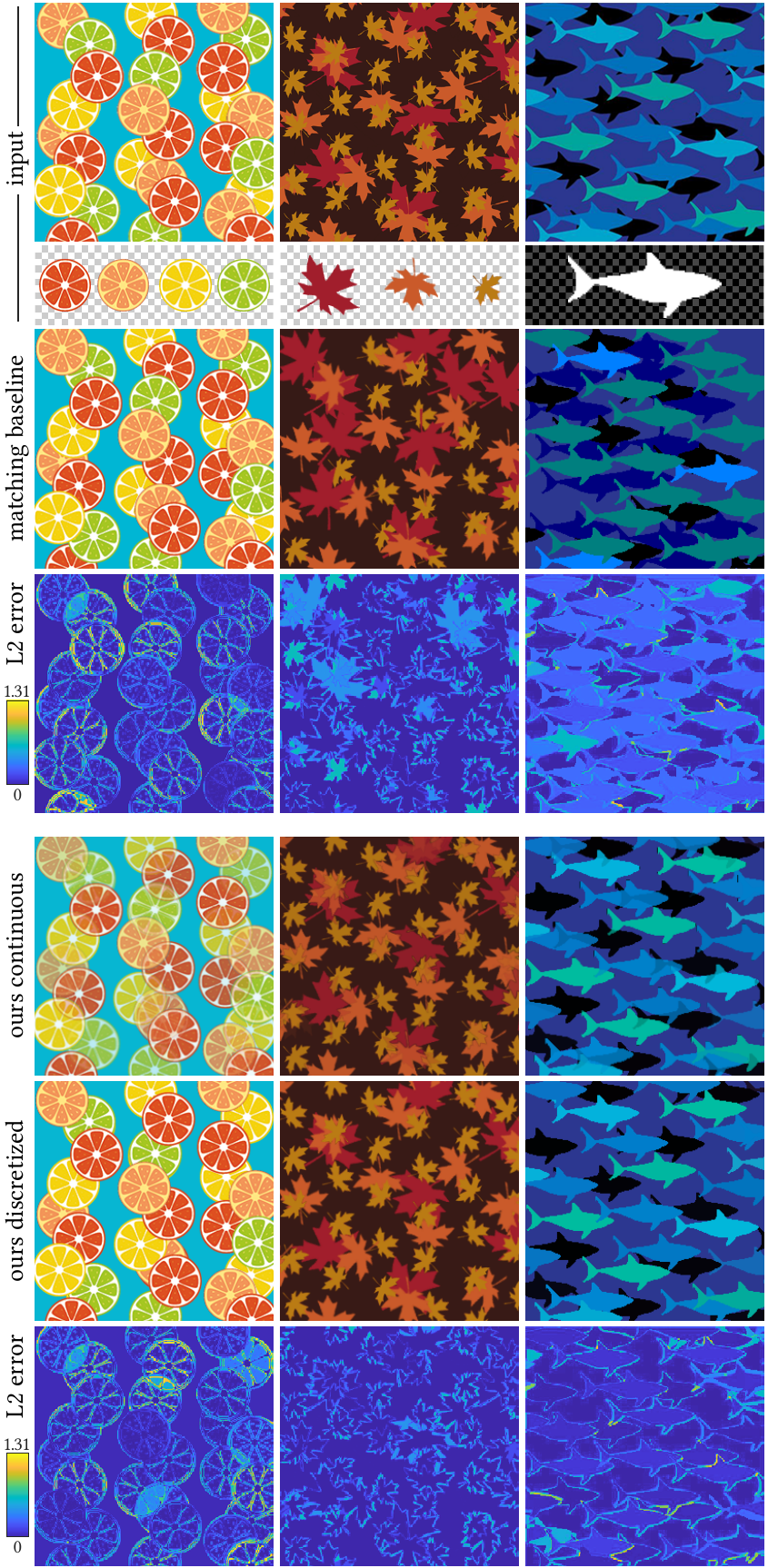}
    \caption{
    {Comparison to a template matching baseline.}  On the left, the results of the baseline are comparable to our approach. When adding more element parameters like orientations in the center, or colors on the right, a dense grid search becomes infeasible, resulting in several false positive and false negative matches. The gradients provided by our compositing function allow optimizing the element positions without requiring a dense grid.}
    \label{fig:results_decomposition_comparison}
\end{figure}

Figure~\ref{fig:results_decomposition_comparison} compares our results to this baseline. As we can expect, in patterns without element rotations (first example), the baseline performs comparable to our approach. In the second example, we also need to optimize over element orientations. However, as we add more parameters, the dimension of the grid increases, making it infeasible to sample the new dimensions densely. Due to inaccuracies in the orientations, we see several errors in the result, including both false positive and false negative matches. This problem worsens as we increase the dimension of the parameter space. In the third example, we optimize over element colors. This adds three dimensions to the parameter space that can only be sampled sparsely to maintain a feasible number of grid points. As a result, we see an increased number of false positives and negatives.

The gradients obtained from our differentiable compositing function, on the other hand, allow for a much sparser sampling of the parameter space, and enable a non-greedy approach where all elements are optimized concurrently.

As second baseline, we compare to the well-known PatchMatch. We match regions in the pattern image $I$ to the image patches in $\mathcal{H}$. More specifically, we run PatchMatch once per image patch $H \in \mathcal{H}$, giving us a set of candidate regions in $I$ that each partially match one of the image patches in $\mathcal{H}$. PatchMatch finds these matching regions starting with a random search strategy, by picking random initial correspondences for each pixel in $I$ to the pixels in $H$. Pixel correspondences are scored based on an L2 distance between small rectangular neigborhoods centered at the corresponding pixels. Good matches are grown into regions. From the resulting set of candidate regions, we filter out regions that have an average matching score below a threshold. For the remaining regions, we obtain the element type $\tau_i$ from image patch $H$ the region is matched to, and the element location $c_i$ from the known correspondences between pixels in the region and pixels in $H$. Since PatchMatch has no notion of layering, we use the matching score of the regions as depth order (highest score is top-most).

\begin{figure}
    \centering
    \includegraphics[width=\linewidth]{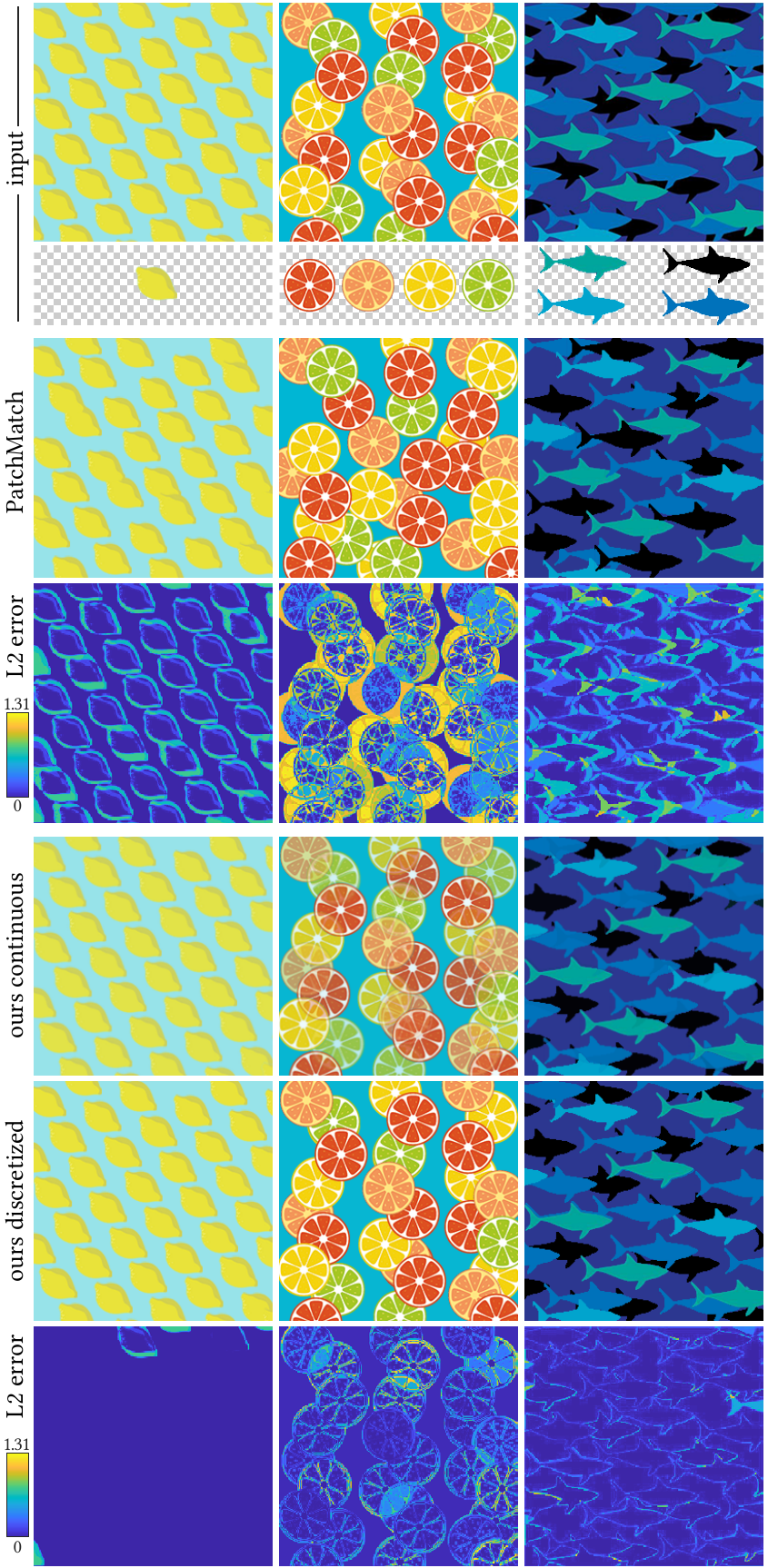}
    \caption{
    {Comparison to PatchMatch.} PatchMatch~\cite{barnes2009patchmatch} uses a random initialization of image-to-patch correspondences, followed by an iterative propagation of good matches. The random search strategy makes finding exact element locations hard, and occlusions further degrade results. Our decomposition uses gradients as guidance, instead of a random search, and explicitly models occlusions, resulting in a more stable decomposition.}
    \label{fig:results_decomposition_comparison_patchmatch}
\end{figure}

Figure~\ref{fig:results_decomposition_comparison_patchmatch} compares our results to PatchMatch. We use patterns without element rotations or color changes, since these would be non-trivial to implement in PatchMatch. PatchMatch works reasonably well on a pattern with simple elements that do not overlap, but fails on more complex patterns with overlapping elements.


\begin{figure}
    \centering
    \includegraphics[width=\linewidth]{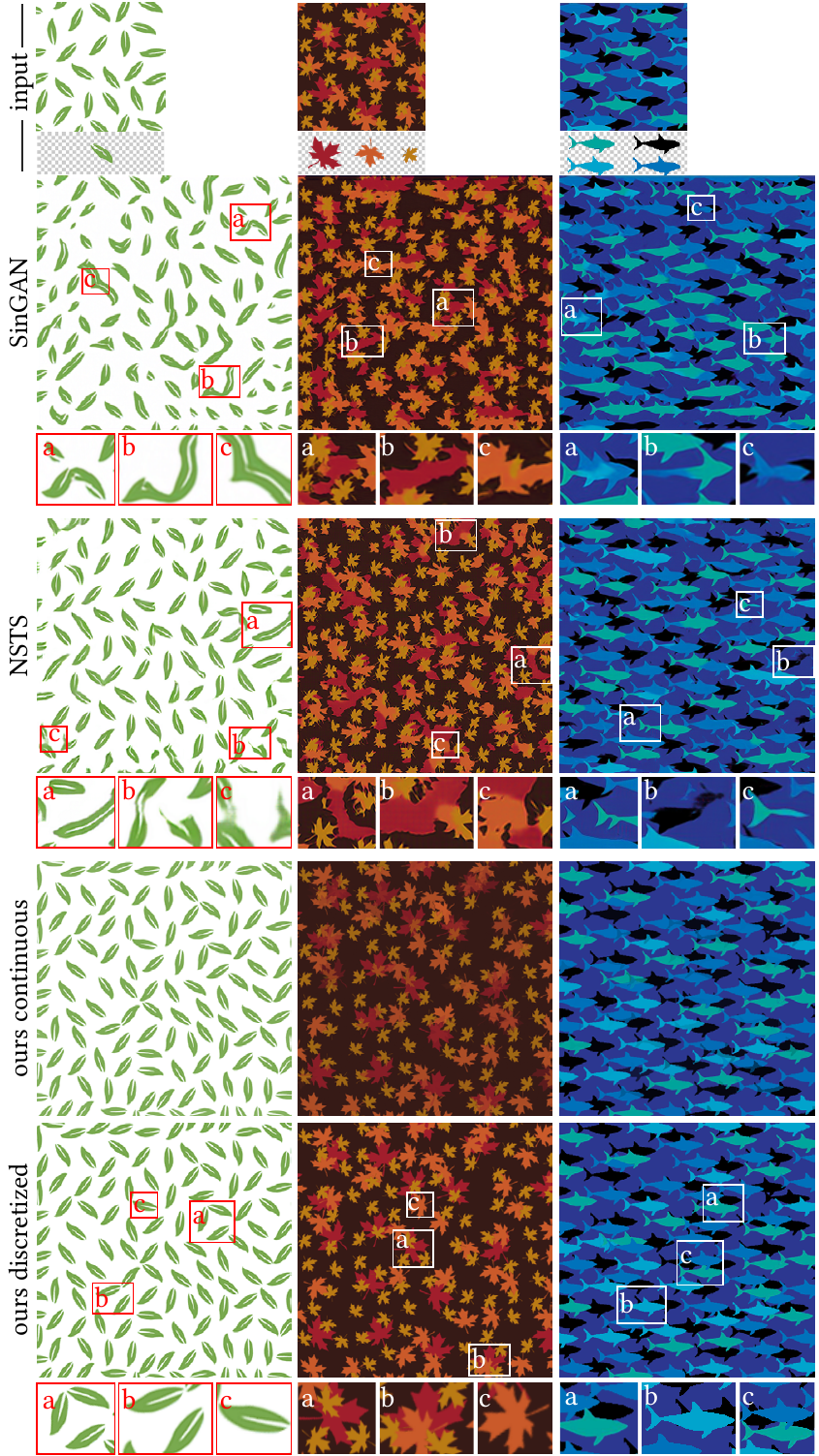}
    \caption{{Pattern expansion.} We expand the pattern image in the top row to twice its size (four times its area) using our differentiable compositor and compare to two state-of-the-art image-based expansion methods, SinGAN~\cite{Shaham_2019_ICCV} and NSTS~\cite{Zhou2018Non-StationaryFormat}. By construction, our results better preserve the integrity of the elements, resulting in a more appealing expansion. Also note that, unlike the image-based methods, the individual elements in our expansion results can be selected and edited.}
    \label{fig:results_expansion}
\end{figure}


\subsection{Pattern Expansion}
\label{sec:results_expansion}

In our second application, our goal is to synthesize a new discrete element pattern that has the same style as a pattern given in an input image, but is generated on a canvas of different size, usually a larger canvas. We measure the difference in style between the composited image $I=\mathcal{F}(\mathcal{E})$ and the input image using a style loss that was originally introduced by Gatys et al.~\shortcite{gatys2016image}:
\begin{equation}
L_s(A, I) := \sum_{l} \frac{w_l}{n_l m_l} \|G^l_A - G^l_I\|_F^2,
\end{equation}
where $G^l_A \in \mathbb{R}^{n_l \times m_l}$ is the Gram matrix of the features in layer $l$ of a pre-trained VGG network~\cite{Simonyan2015VERYRECOGNITION} that was applied to image $A$ and $w_l$ are per-layer weights. See the original work of Gatys et al. for details. We use layers $2$, $5$, $10$, and $15$ of a pre-trained VGG-19 network with all weights $w_l = 0.2$. Note that the Gram matrices represent feature statistics that remain comparable across different image sizes.
The optimal elements $\mathcal{E}^*$ are the minimum of this loss:
\begin{equation}
    \mathcal{E}^* := \argmin_{\mathcal{E}} L_s(A, \mathcal{F}(\mathcal{E})).
\end{equation}

\paragraph{Workflow}
Similar to pattern decomposition, the user first selects one example of each element type from the input image. Then, a canvas size can be chosen that determines the size of the expanded pattern. After optimization, the user obtains a pattern in the chosen canvas that has a style similar to the input image and is composed of discrete elements.

\paragraph{Baselines}
To show the advantages of generating a pattern with discrete elements, we compare against two state-of-the-art image-based pattern synthesis methods.
SinGAN~\cite{Shaham_2019_ICCV} trains a GAN on a single image. Their model learns to generate images with similar patch statistics as the input image. This is enabled by a loss similar to the style loss we are using, but replaces the Gram matrix with a manually defined feature statistic with a learned discriminator.
As a second baseline, we use the Non-stationary Texture Synthesis (NSTS) method by Zhou et al.~\shortcite{Zhou2018Non-StationaryFormat}. This method also trains a generator with an adversarial loss on a single input image and additionally uses the style loss from Gatys et al.~\shortcite{gatys2016image}.

While the style losses are slightly more advanced in these baselines, we focus on comparing our use of discrete elements to the baselines' approach of directly generating an image. An interesting direction for future work lies in using our differentiable function $\mathcal{F}$ as a differentiable component in a network like SinGAN or NSTS to output a set of discrete elements instead of an image.

Additionally, we compare to a recent Point Pattern Synthesis (PPS) method by Tu et al.~\shortcite{tu2019point}. This method aims at expanding patterns of 2d points in a plane to a larger canvas. The points are rendered as small Gaussian hats to an image, where an image-based style loss is optimized to get an expanded image. The expanded image is then converted back to points. The authors use a histogram- and correlation-based loss in addition to a style loss similar to $L_s$. The baseline supports multiple point types by coloring the rendered Gaussians, but does not support orientations or layering.
We use the centers of our elements as points, color them by element type using well-separated colors, use a random occlusion order, and random orientations if the pattern has elements with multiple orientations.

\begin{figure}
    \centering
    \includegraphics[width=\linewidth]{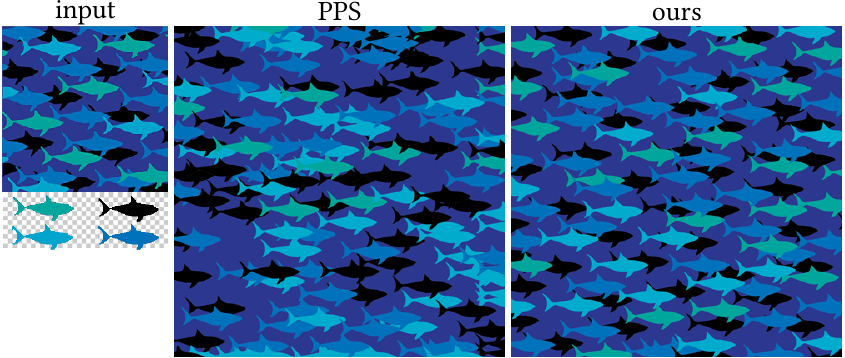}
    \caption{{Comparison to point-based pattern synthesis.} We compare a point-based pattern synthesis method to our approach. Note how the baseline preserves the layout of the input pattern less accurately than our method. Unlike the point-based method, our method can take into account properties of the pattern that depend on the shape of the elements, such as amounts of overlap or gap sizes between elements.}
    \label{fig:results_expansion_pps}
\end{figure}

\paragraph{Results}
Figure~\ref{fig:results_expansion} compares our expansion results to the two image-based baselines. We show a simple example with a single element type and no overlaps to clearly illustrate typical errors of the image-based methods, and two more complex examples with multiple elements and overlaps. In all examples, we can see that the image-based methods cannot preserve the integrity of elements, since they do not have a notion of discrete elements. Separate elements are merged together and the shapes of individual elements are not maintained. Even though the style loss we are using for our experiments is less advanced than the baselines, we still obtain more appealing results, due to our use of discrete elements.

We compare with an expansion result of the PPS baseline in Figure~\ref{fig:results_expansion_pps}. We can see that the method struggles to synthesize a pattern with a similar layout as the input pattern. This baseline is designed for tightly packed patterns of points and performance drops on our larger, more spread out elements. In our patterns, knowledge about the shape of the elements is necessary to obtain cues about important properties of the pattern like amounts of overlap and typical gap sizes between neighboring elements. Additionally, this baseline does not support layering.
The output of our method, on the other hand, does take into account the shape of elements, giving the style loss more to work with, and allowing us to synthesize a stylistically correct layering.

\begin{figure}
    \centering
    \includegraphics[width=\linewidth]{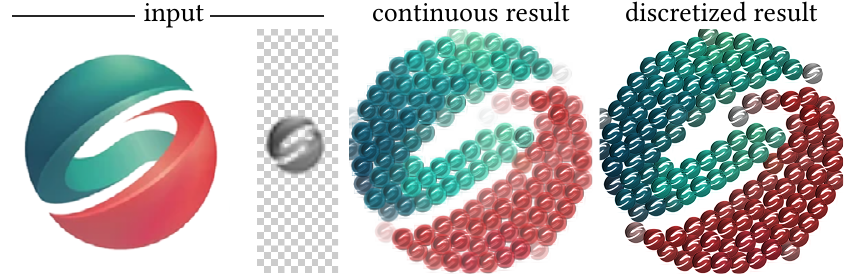}
    \caption{{Generating mosaics.} By decomposing a non-pattern image and adding a loss that penalizes overlaps between elements, we can create mosaics. Here we create a mosaic of the Siggraph Asia logo.}
    \label{fig:results_mosaic}
\end{figure}

\subsection{Additional Applications}
\label{sec:results_additional}
Our differentiable compositing function can be used as a component in a large range of applications. Here we show a few additional examples.

\paragraph{Making patterns tileable.}
By combining the $L_2$ loss and the style loss, we can make an existing pattern image tileable, as shown in Figure~\ref{fig:teaser}. We first decompose the pattern as described in Section~\ref{sec:results_decomp}. This gives us discrete elements that can be edited manually. To make the pattern tileable horizontally, we can, for example, delete the elements that are clipped on the right border, select the elements that are clipped on left border and copy them over to the right border (i.e., translate the copies in x direction by the width of the image). A similar approach can be used to make the pattern tileable vertically. This manual operation however, might compromise the style of the pattern; it could, for example, create incorrect overlaps between elements, create holes, or introduce incorrect spacing. To restore the style of the pattern, we optimize the elements in the edited pattern to have the same style as the original pattern image using the loss $L_s$. We initialize the elements to their current positions and keep the elements near the border that ensure editability fixed during the optimization. This results in a tileable pattern with the same style as the input pattern.

\paragraph{Generating mosaics.}
While our compositing function can only obtain an exact reconstruction of images that are composed of a few element types, there is nothing stopping us from minimizing the loss $L_d$ on arbitrary images. This creates a mosaic that reconstructs the input image with the given set of element types. Since mosaics typically have non-overlapping elements, we introduce an auxiliary loss that penalizes overlaps: $L_o =\frac{1}{P} \sum_{p=1}^P \max\big(0, -1 + \sum_k M_k(x_p)\big)$, where we sum the occupancy masks $M_k$ of all layers, and any values larger than 1, indicating overlaps, are penalized. The outer sum takes the average over all pixels in the occupancy masks.
An example is shown in Figure~\ref{fig:results_mosaic}, where create a mosaic of the Siggraph Asia logo with small Siggraph Asia logos.

\subsection{Limitations}
\begin{figure}[h!]
    \centering
    \includegraphics[width=\linewidth]{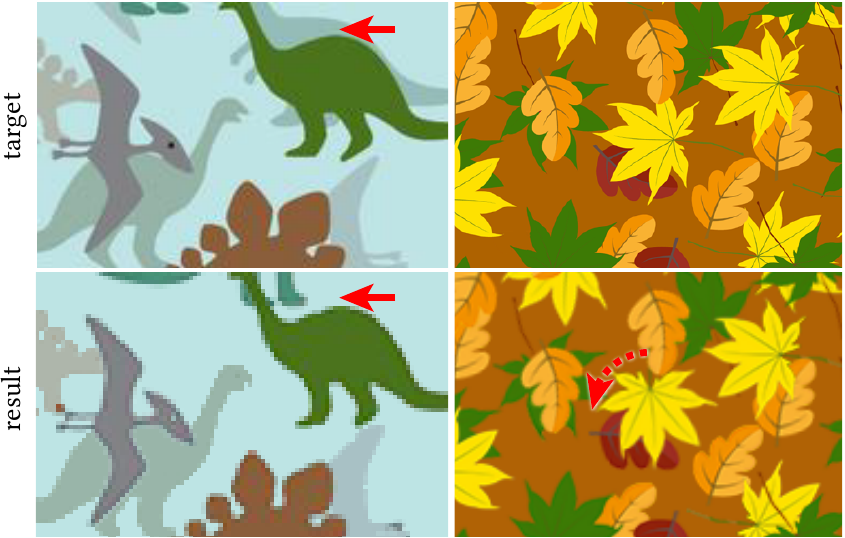}
    \caption{
    {Failure Cases.} 
    Optimization results can be deteriorated by faint elements (column 1) or elements that are close to rotationally symmetric (column 2).}
    \label{fig:failure_cases}
\end{figure}

One limitation of our method is that the output domain of our compositing function is restricted to patterns that are composed from multiple instances of a small set of atomic image patches. This includes many types of patterns found on the web, but also excludes some types, like patterns with continuously changing shapes, for example a pattern containing of continuous deformations between a circle and a rectangle. 
A second limitation is that users have to mark one instance of each element type in advance. This is still far easier than marking all elements, but requires some manual work. Also optimizing over the image patches $\mathcal{H}$ would remove the need for this manual input and is an interesting avenue for future work.
Finally, our optimization is not always guaranteed to find the global optimum. A few typical failure modes of our optimizations are shown in Figure~\ref{fig:failure_cases}, columns 1 and 2. Faint elements, or elements very only few pixels are visible give small gradient magnitudes resulting in missed elements (column 1). Elements that close to rotationally symmetric (column 2) cause strong local minima in the space of orientations, resulting in elements with incorrect orientations.

\section{Conclusion and Future Work}
We presented a differentiable composting function for discrete element patterns.
We have shown that the gradients provided by this function can benefit pattern editing workflows in several applications.
First, our differentiable compositor can be used to decompose a pattern image into its discrete elements. Unlike existing template matching approaches, our gradients allow us to handle a higher-dimensional parameter space for elements, such as element orientations and colors.
Second, we can use our compositor to expand an existing pattern on a larger canvas by minimizing a style loss. Unlike current image-based approaches, our focus on discrete elements guarantees the integrity of elements in the expanded pattern and leads to more appealing expansion results.
In addition to these two main applications, we demonstrate the versatility of our method on two more specialized applications.

In future work we would like to incorporate our differentiable element compositor as a component in generative modeling approaches. For example, using our compositor in the generator of a GAN would give us the benefits in quality that a GAN provides, while our compositor would guarantee the integrity of individual elements, and make the resulting pattern editable by producing individual elements and their layering as output.

There are also several avenues to improve the differentiable function itself.
First, it would be interesting to solve for the image patches $\mathcal{H}$ in addition to element properties $\mathcal{E}$. This would remove the need to manually specify the element types and open up a wider range of new applications.
Second, defining a depth value per pixel of an element's occupancy mask instead of per element would allow handling non-planar element surfaces and enable local-layering among elements (cf., \cite{locallayering:09}).
Finally, since our gradients allow us to handle high-dimensional element parameters, the range of patterns our approach can handle could be extended by adding additional element parameters, such as scale or transparency.



\begin{acks}

We would like to thank the anonymous reviewers for their helpful suggestions and Maks Ovsjanikov for discussions in an early phase of this project. 
This research was supported by an ERC Grant (SmartGeometry 335373), Google Faculty Award, and gifts from Adobe.
\end{acks}

\bibliographystyle{ACM-Reference-Format}
\bibliography{extra.bib}

\end{document}